\documentclass[preprint,10pt]{elsarticle}

\usepackage{floatrow}
\newfloatcommand{capbtabbox}{table}[][\FBwidth]
\usepackage{multirow}
\usepackage{epstopdf}
\usepackage{subfigure}
\usepackage{geometry}
\usepackage{lineno}
\usepackage{amssymb}
\usepackage{amsthm, amsmath,amssymb,amsfonts}
\theoremstyle{plain}

\theoremstyle{definition}

\usepackage{multirow}
\usepackage{pifont}
\newtheorem{ass}{Assumption}
\newcommand{\numstd}[2]{
    #1{\scriptsize$\pm$#2}
}

\usepackage{algorithmic}
\usepackage{graphicx}
\usepackage{textcomp}
\usepackage{booktabs}
\usepackage[table,xcdraw]{xcolor}
\usepackage[colorlinks, citecolor=black]{hyperref}
\definecolor{mycolor}{RGB}{218, 242, 251}
\AtBeginDocument{%
}

\begin{document}

\begin{frontmatter}



\title{LOCAL: Learning with Orientation Matrix to Infer Causal Structure from Time Series Data}

\author{Jiajun Zhang$^a$, Boyang Qiang$^b$, Xiaoyu Guo$^a$, Weiwei Xing\thanks{1}$^{a}$ , Yue Cheng\thanks{1}$^{a}$, Witold Pedrycz$^c$}

\address{$^a$ School of Software Engineering, Beijing Jiaotong University, Beijing, P.R.China}
\address{$^b$ Department of Computer Science, Hong Kong Baptist University, Hong Kong, P.R. China}
\address{$^c$ Department of Electrical and Computer Engineering,
University of Alberta, Edmonton, Canada}

\tnotetext[label1]{WeiWei Xing and Yue Cheng are the Co-Corresponding authors. E-mail: wwxing@bjtu.edu.cn, yuecheng@bjtu.edu.cn}

\begin{abstract} Discovering the underlying \textbf{\textit{Directed Acyclic Graph}} (DAG) from time series observational data is highly challenging due to the dynamic nature and complex nonlinear interactions between variables. Existing methods typically search for the optimal DAG by optimizing an objective function but face scalability challenges, as their computational demands grow exponentially with the dimensional expansion of variables. To this end, we propose \textbf{LOCAL}, a highly efficient, easy-to-implement, and constraint-free method for recovering dynamic causal structures. \textbf{LOCAL} is the first attempt to formulate a \textbf{\textit{quasi-maximum likelihood-based}} score function for learning the dynamic DAG equivalent to the ground truth. Building on this, we introduce two adaptive modules that enhance the algebraic characterization of acyclicity: \textbf{\textit{Asymptotic Causal Mask Learning}} (ACML) and \textbf{\textit{Dynamic Graph Parameter Learning}} (DGPL). ACML constructs causal masks using learnable priority vectors and the Gumbel-Sigmoid function, ensuring DAG formation while optimizing computational efficiency. DGPL transforms causal learning into decomposed matrix products, capturing dynamic causal structure in high-dimensional data and improving interpretability. Extensive experiments on synthetic and real-world datasets demonstrate that \textbf{LOCAL} significantly outperforms existing methods and highlight \textbf{LOCAL}'s potential as a robust and efficient method for dynamic causal discovery.
\end{abstract}
\begin{keyword}Causal structure, Time series, Directed acyclic graph, Constraint free
	
	
	
\end{keyword}

\end{frontmatter}

\begin{figure}[tb!]
    \centering
    \includegraphics[width=1\linewidth]{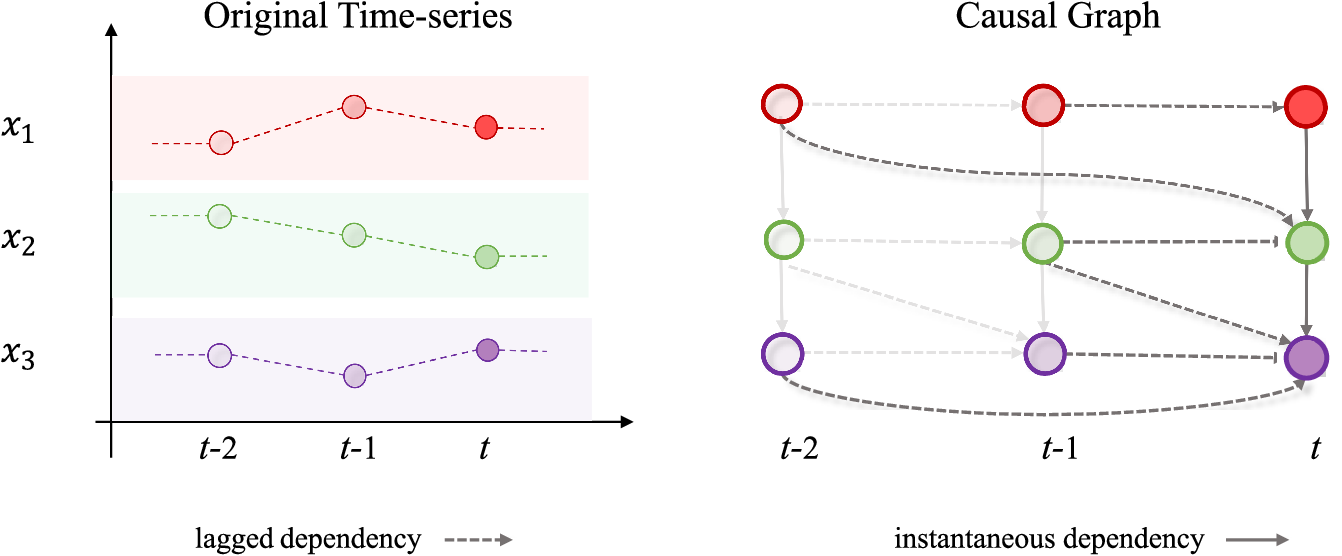}
    \caption{Illustration of instantaneous dependencies (solid lines) and lagged dependencies (dashed lines) in a dynamic Bayesian network (DBN) with $d = 3$ nodes and an autoregression order of $p = 2$. For clarity, edges that do not influence the variables at time $t$ are shown in a lighter shade.}
    \label{fig:enter-label}
\end{figure}

Exploration of the underlying causal generation process of dynamic systems is an important task \citep{Cheng_2024, 10.1145/3616855.3635766} for trustworthy machine learning. Unfortunately, it is unethical, impossible due to technical reasons, or expensive to conduct intervention experiments on the dynamic systems of certain domains \citep{Adam2023, 10.1145/3539597.3570461}. Another challenge is to infer about the structure which may be high dimensional and nonlinear. Some recent works \citep{Pamfil20a, Sun23c, Gao22a, Fan2023} have made significant efforts by employing \textit{dynamic Bayesian networks} (DBNs) with observational and interventional data: among dynamic systems, as illustrated in Figure \ref{fig:enter-label}. The variable $x_i$ at timestep $t$ is affected by which variables $x_j$ at the same time step (instantaneous dependency) and which variables $x_j$ at the previous timestep (lagged dependency)? This question highlights the crucial roles of those algorithms in the interpretable performance of the trained models. 

In order to study the nonparametric DBN, DYNO \citep{Pamfil20a} (\emph{i.e.}, a score-based approach to learning DBNs with a differentiable DAG constraint \citep{Zheng2018}) was proposed as a proxy to capture the parents of child variables. However, in Section \ref{Section:simulation}, our practical analysis shows that the DYNO algorithm and its extensions \citep{Sun23c, Gao22a, Fan2023} adopting matrix exponential constraints require an extremely long time to optimize in high-dimensional dynamic systems, even if they smartly adopt interventional data to enhance the identifiability \citep{Li2023, Li2024}.  Then, it is natural to need a proxy model that can infer dynamic causal structures in high-dimensional situations faster, which is also the main goal of our work.

Recently, much of the research on dynamic causal structure learning has concentrated on applying soft sparsity and DAG constraints. For instance, Golem \citep{Golem_2020_nips} formulated a likelihood-based score function for handling the causality of thousands of nodes. Yu et al. \citep{10230429} extended it for the recovery of dynamic causal structure. Concurrently, \citep{Fang2023} verified the feasibility of further accelerating the learning of DAG based on this likelihood function both theoretically and experimentally. These approaches are aimed at enhancing flexibility and scalability in high dimensions and circumventing rigid structural constraints.

Building on these insights,  we propose a novel framework for \textbf{\underline{L}}earning with \textbf{\underline{O}}rientation matrix to infer \textbf{\underline{CA}}usa\textbf{\underline{L}} structure from time series data, which we call \textbf{LOCAL}. We develop a quasi-maximum likelihood-based dynamic structure learning method with identifiability guarantee. Powered by this \textbf{\textit{quasi-maximum likelihood-based objective}}, we propose to enhance the algebraic characterization of acyclicity with two adaptive modules for causal structure recovering task: 1) an Asymptotic Causal Mask Learning (ACML) module which leverages learnable priority vectors ($\boldsymbol p$) and the Gumbel-Sigmoid function to generate causal masks, ensuring the creation of directed acyclic graphs (DAGs) while optimizing computational efficiency; 2) a Dynamic Graph Parameter Learning (DGPL) module to transform causal learning into decomposed matrix products ($\boldsymbol W = \boldsymbol E_s \boldsymbol E^T_t$), effectively capturing the dynamic causal structure of high-dimensional data and enhancing interpretability. Those leads us to exploit faster gradient-based optimization, such as Adam \citep{KingBa15}, and GPU acceleration.

\noindent \textbf{Contribution.} The main contributions of this work are as follows:
\begin{itemize}
    \item To the best of our knowledge, this work represents the \textit{first} principled formulation of a quasi-maximum likelihood score function for dynamic causal structure learning. Our theoretical analysis and empirical validation demonstrate its superior robustness and accuracy compared to existing approaches.
    
    \item We propose \textbf{LOCAL}, an innovative framework that integrates two adaptive modules, \textbf{ACML} and \textbf{DGPL}, effectively eliminating the need for matrix exponential operations in causal learning.
    
    \item We conduct extensive experiments on both synthetic datasets and real-world benchmarks. On a synthetic dataset with 100 nodes, our method achieves impressive instantaneous and lagged F1 scores of 0.81 and 0.59, respectively.
\end{itemize}

The remainder of this paper is structured as follows. In Section.~\ref{sec:work}, we discuss the challenges of dynamic causal discovery and the limitations of existing methods. In Section.~\ref{sec:local}, we introduce our quasi-maximum likelihood-based objective function and detail the design of the ACML and DGPL modules for learning dynamic DAG structures and improving nonlinear modeling. In Section.~\ref{sec:exp}, we present experimental results demonstrating the advantages of LOCAL in accuracy, efficiency, and interpretability across synthetic, NetSim, and real-world datasets. We also perform ablation studies and parameter analysis. Finally, in Section.~\ref{sec:conclusion} and Section.~\ref{sec:limitation}, we summarize efficiency of LOCAL and discuss potential future extensions.

\section{Related Work}
\label{sec:work}
\subsection{Score-Based Objective for Causal Discovery}
 Learning causal structure from data using constraint and score-based techniques was a classic approach \citep{maxwell1997efficient}, but the major drawback was the computational inefficiency. Many recent works have contributed to promoting computability and practicality: NOTEARS \citep{Zheng2018} employed a least squares objective to speed up the process but required an augmented Lagrangian to satisfy hard DAG constraints. Based on this, DYNO \citep{Pamfil20a} first tried to recover the causal structure of time series data. Golem \citep{Golem_2020_nips} exploited the likelihood-based function to make the objective unconstrained. Those likelihood-based works, however, still struggled with the DAG constraint.

\subsection{Constraint Free Causal Discovery}
Many efforts have been made to emancipate constraints DAGs, focusing on designing more tractable optimization constraints \citep{Yu19a, Bello2022, cai2023recovering, Ferdous23a} on static data. In contrast to these approaches, Yu et al. \citep{Yu2021} introduced NOCURL, which separated the topological ordering from the adjacency matrix, thereby eliminating the necessity to optimize the DAG constraint within the objective function. Inspired by NOCURL, a series of causal masks based on topological structures appeared \citep{ng2022, Annadani2023, Massidda2024}.  For example,  MCSL-MLP \citep{ng2022} sampled DAGs from a real matrix $U \in \mathbb{R}^{d \times d}$. BayesDAG \citep{Annadani2023} inherited NOCRUL's grad matrix and directly sampled DAGs from the posterior of grad matrix $\boldsymbol p(u, v)$. By exploiting the temperature sigmoid function to $\boldsymbol p(u, v)$, COSMO \citep{Massidda2024} provided a fast implementation method when sampled DAGs. However, DAG masks for the time domain datasets have rarely been studied.

\subsection{Matrix Decomposition for Causal Discovery} 
The matrix decomposition approach was widely employed to address high-dimensional problems \citep{Fang2023, Massidda2024, CHEN2024111868}, with applications spanning large-scale parameter tuning \citep{hu2022lora}, image restoration \citep{Chen2022Lowrank}, and trajectory prediction \citep{Bae_2023_ICCV}. While many of these works focused on learning intrinsic dimensionality to achieve efficient optimization in parameter space, VI-DP-DAG \citep{Charpentier2022} adopted adjacency matrix decomposition $\boldsymbol{W}$ as $\boldsymbol{W} = \amalg^T U \amalg$, while LoRAM \citep{Dong2023} integrated low-rank matrix factorization with sparsification mechanisms for continuous optimization of DAGs. Those works did not conduct an in-depth exploration of the physical meaning of the decomposition matrix, and the decomposition form was not concise enough.


\section{LOCAL: Learning Causal Discovery from Time Series Data}
\label{sec:local}
\subsection{Problem Definition}
We target the problem of causal discovery from a dynamic system. Consider multidimensional stationary time series that contains $d$ causal univariate time series represented as $\mathcal{X}:= [X^i_{t}] \in \mathbb{R}^{T \times d}$, where $X^i_{t}$ is the recording of $i$-th component of $X_{t}$ at time step $t$. Our target is to simultaneously infer the \textbf{\textit{instantaneous dependency}} $\boldsymbol{W}^i$ = \{$\boldsymbol{W}^{i,j}: X^i_{t} \rightarrow X^j_{t} | i,j \in 1,\cdots, d; i \neq j\}$ and the \textbf{\textit{lagged dependency}} $\boldsymbol{A}^{i}$  = \{$\boldsymbol{A}_k^{i,j}: X^j_{t-p} \rightarrow X^i_{t} | i,j \in 1,\cdots, d; p \in Z^{+} \}$ of $X^i_{t}$, from which we may recovery the dynamic causal structure. Following the practice in the dynamic causal structure discovery, we formulate the problem of learning a \textbf{\textit{dynamic Bayesian network} (DBN)} that captures the dependencies in the $X^i_{t}$:

\begin{equation}
\begin{gathered}
    X_t = X_t \boldsymbol{W} + Y\boldsymbol{A} + N \\
    s.t. \text{ } h(\boldsymbol{W})=0
\end{gathered}
\label{eq1}
\end{equation}
where $\boldsymbol{Y} = [{X_{t-1}}|\cdots|{X_{t-p}}] \in \mathbb {R}^{1 \times pd}$ is the $p$ order time-lagged version of  ${X_t} \in \mathbb {R}^{1 \times d}$, and $\boldsymbol{{A}}=[{A}_1| \cdots| {A}_p] \in \mathbb {R}^{pd \times d}$; $N$ is the exogenous noise variables that are jointly Gaussian and independent; $h(\boldsymbol{W})= Tr(e^{\boldsymbol{W} \circ \boldsymbol{W}}-d)$ is a differentiable equality DAG constraint. Here, ''$\circ$" denotes the Hadamard product of two matrices. 
\subsection{Quasi-Maximum Likelihood-Based Objective}
Due to the hard acyclicity constraint \citep{Pamfil20a}, learning a DAG equivalent to the ground truth DAG through directly optimizing Eq. (\ref{eq1}) is difficult. To circumvent this issue, we reformulate a score-based method to maximize the data quasi-likelihood of a dynamic linear Gaussian model.

The overall objective is defined as follows:

\begin{equation}
\begin{gathered}
    \min_{\boldsymbol{W}, \boldsymbol{A}} \mathcal{S}(\boldsymbol{W}, \boldsymbol{A}; X_t) = \mathcal{L}(\boldsymbol{W}, \boldsymbol{A}; X_t) \\
    + \lambda_1 h(\boldsymbol{W}) + \lambda_2 (\mathcal{R}_{sparse}(\boldsymbol{A}) + \mathcal{R}_{sparse}(\boldsymbol{W}))
\end{gathered}
\label{eq2}
\end{equation}
where $\mathcal{L}(\boldsymbol{W}, \boldsymbol{A}; X_t)$ is the quasi-maximum likelihood estimator (QMLE) of $\mathcal{X}$, $\mathcal{R}_{sparse}$ is the penalty term encouraging sparsity, \emph{i.e.} having fewer edges.

\begin{ass}
    Suppose $N$ follows the multivariate normal distribution $\mathcal{N}(0, \sigma_i^2I), \, i = 1, 2, \dots, d$. Further assumes that the noise variances are equal, \emph{i.e.} $\sigma_1^2 = \cdots = \sigma_d^2 = \sigma^2$. Let $X^{-i}_t = (X^{j}_t, j \neq i)^T$ denote the responses of all variables except for the $i$-th variable.  Motivated by the  least squares estimation (LSE) method \citep{ZHU2020591}, we have $\tilde{X}^i_t = E \{ X^i_t | X^{-i}_t\} = \mu_i + \sum_{j \neq i} \alpha_{i,j} (X^j_t - \mu_j)$, where

    \begin{equation}
            \alpha_{ij} = \frac{(w_{ij}+w_{ji})-\sum_{k}w_{ki}w_{kj}}{1+ \sum_{k}w^2_{ki}}
            \label{3}
    \end{equation}
and $\mu_i = E(X^i_t)$. 
\end{ass}

Inspecting Eq. (\ref{3}), we can find that for the $i$-th variable, the causal relationship are related to its first- and second-order parents or children, where the first-order parents or children are collected by $\{j: w_{ij} \neq 0 \text{ or } w_{ji} \neq 0 \}$, and the second-order parents or children are collected by $\{j: \sum_{k}w_{ki}w_{kj} \neq 0\}$. Based on the conditional expectation and Eq. (\ref{3}), we now derive that:

\begin{equation}
\begin{aligned}
    \tilde{X}^i_t - \mu_i &= (1+\|\boldsymbol{W}_{\cdot i}\|^2)^{-1}(W_{\cdot i} + \boldsymbol{W}^T_{\cdot i} - \boldsymbol{W}^T_{\cdot i}W_{ii})(x_t-\mu) \\
      & = (1+\|\boldsymbol{W}_{\cdot i}\|^2)^{-1} \{(\boldsymbol{W}_{\cdot i} + \boldsymbol{W}^T_{\cdot i} - \boldsymbol{W}^T_{\cdot i}W_{ii})(x_t-\mu) \\
      & - ({X}^i_t - \mu_i)\} + ({X}^i_t - \mu_i)
  \end{aligned}
  \label{eq4}
\end{equation}

By defining $\boldsymbol{D} := \{\boldsymbol{I} + diag(\boldsymbol{W}^T\boldsymbol{W})\}^{-1}$ and $\boldsymbol{S}:=\boldsymbol{I}-\boldsymbol{W}$, we can verify the least squares objective function as

\begin{equation}
    \frac{1}{n} \sum^d_{i=1}\|X^i_t \boldsymbol{S} - Y^i\boldsymbol{A}\|^2 = \frac{1}{n} \| \{X_t\boldsymbol{S} - Y\boldsymbol{A}\} \boldsymbol{D}\boldsymbol{S}^T\|^2
    \label{eq5}
\end{equation}

Following the multivariate Gaussian distribution assumption, we now formulate a score-based method to maximize the data quasi-likelihood of the parameters in Eq. (\ref{eq2}). Omitting additive constants, the log-likelihood is:

\begin{equation}
\begin{aligned}
\mathcal{L}(\boldsymbol{W}, \boldsymbol{A}; X_t) &= \frac{1}{2} \log \sigma^2 - \log|\text{det}(\boldsymbol{S})| 
    - \frac{1}{2} \frac{(X_t \boldsymbol{S}- Y\boldsymbol{A})' (X_t\boldsymbol{S} - Y\boldsymbol{A})}{\sigma^2}
\end{aligned}
\label{eq6}
\end{equation}


 The QMLE in most common use is the maximizer of $\mathcal{L}(\boldsymbol{W}, \boldsymbol{A}; X_t)$. Solving $\frac{\partial \mathcal{L}}{\partial \sigma^2} = 0$ yields the estimate,

\begin{equation}
\hat{\sigma}^2 = \frac{1}{n} [X_t\boldsymbol{S} - Y\boldsymbol{A}]'[X_t\boldsymbol{S} - Y\boldsymbol{A}]
\label{eq7}
\end{equation}

Maximization w.r.t. $\sigma^2$ gives $\hat{\sigma}^2 := \frac{1}{n} [X_t\boldsymbol{S} - Y\boldsymbol{A}]'[X_t\boldsymbol{S} - Y\boldsymbol{A}]$, the corresponding MLE is, again omitting additive constants,
\begin{equation}
    \mathcal{L}(\boldsymbol{W}, \boldsymbol{A}; X_t) = \frac{d}{2} \log ( \|X_t\boldsymbol{S} - Y\boldsymbol{A}\|^2) - \log|\text{det}(\boldsymbol{S})|
\label{eq8}
\end{equation}

Note that the $\|X_t\boldsymbol{S} - Y\boldsymbol{A}\|^2$ term in Eq. (\ref{eq8}) is the least squares objective, replace it with Eq. (\ref{eq5}), and we could verify the objective function as

\begin{equation}
\begin{gathered}
\mathcal{L}(\boldsymbol{W}, \boldsymbol{A}; X_t) = \frac{d}{2} \log ( \| \{X_t\boldsymbol{S} - Y\boldsymbol{A}\}\boldsymbol{D}\boldsymbol{S}^T \|^2) - \log|\text{det}(\boldsymbol{S})|
\end{gathered}
\label{eq9}
\end{equation}

\subsection{Asymptotic Causal Mask Learning}
Although QMLE can be statistically efficient, the computational cost could be expensive due to the matrix exponential operation of $h(\cdot)$ function \citep{Zheng2018}.  Most recent work in causal structure learning deploys an alternative continuous DAG constraint $h(\boldsymbol{W})$ to characterize acyclic graphs. 
Decoupling the causal structure into a causal mask and a real matrix can improve explainability for instantaneous causal effects. However, this requires the matrix to adapt from $d \times d$ to $d \times (d+1)$. To address this issue, we propose the \textbf{\textit{Asymptotic Causal Mask Learning} (ACML)} module , which asymptotically infers causal relationships from data. The ACML module initializes a learnable priority vector $\boldsymbol{p} \in \mathbb{R}^d$, which is initialized as a vector with all elements set to 1, and a strictly positive threshold $\omega > 0$.  The smooth orientation matrix of the strict partial order is given by the binary orientation matrix $\boldsymbol{M}_{\omega}(\boldsymbol{p})_{u,v} \in \{0, 1\}^{d \times d}$ such that:

\begin{equation}
\begin{gathered}
    \boldsymbol{M}_{\tau, \omega}(\boldsymbol{p})_{u,v} = \sigma_{\tau, \omega}(\boldsymbol{p}_{v}-\boldsymbol{p}_{u})
\end{gathered}
\end{equation}
where $\sigma_{\omega}$ can be formulated as the $\omega$-centered Gumbel-Sigmoid:
\begin{equation}
\sigma_{\tau, \omega} (\boldsymbol{p}_{v}-\boldsymbol{p}_{u}) = \frac{exp((\boldsymbol{p}_{v}-\boldsymbol{p}_{u}-\omega) + \dot{g})/ \tau}{(exp((\boldsymbol{p}_{v}-\boldsymbol{p}_{u}-\omega) + \dot{g})/ \tau)+ exp(\ddot{g}/ \tau)}
\end{equation}
where $\sigma_{\omega}$ is formulated as the $\omega$-centered Gumbel-Sigmoid, $\dot{g}$ and $\ddot{g}$ are two independent Gumbel noises, and $\tau \in (0, \infty)$ is a temperature parameter. The threshold $\omega$ adjusts the midpoint of the sigmoid function and disrupts the symmetry when two variables have roughly equal priority.  Finally, the ACML enhanced DBN can be formulated as:

\begin{equation}
\begin{gathered}
    X_t = X_t (\boldsymbol{W} \circ \boldsymbol{M}_{\tau, \omega}(\boldsymbol{p})_{u,v}) + Y\boldsymbol{A} + Z\\
\end{gathered}
\label{eq10}
\end{equation}

\subsection{Dynamic Graph Parameter Learning}
Another problem lies in the existing dynamic causal structure models, which require directly learning $p+1$ $d \times d$-dimensional matrices. There are two main methods to capture dynamic causal effects: 1) Designing a complexity neural network, which specifically customizes a neural network with DAG constraint \citep{Gao22a, Misiakos2023}. 2) With the help of the Granger causal mechanism, \citep{Sun23c} introduces the idea of Neural Granger's component network \citep{Tank2021} to provide each variable design independent neural network. However, these approaches are quite counter-intuitive. Overly complex network design or component Neural Granger \citep{cheng2024causaltime, zhou2024jacobian} methods are overparameterized, but most works in practice only exploit the first-layer weights of the network to characterize causal effects, which increases the optimization cost and has to adopt a second-order optimizer L-BFGS-B. Besides, over-parameterization methods are prone to over-fitting when dealing with unsupervised problems such as causal structure discovery, leading to considerable biases in the causal structure.

To solve the issue, we propose a \textbf{\textit{Dynamic Graph Parameter Learning} (DGPL)} module to infer the dynamic causal graphs from observed data and interventional data automatically. The DGPL module first randomly initializes 2 learnable dynamic node embedding dictionaries $\boldsymbol E_{so} \in \mathbb{R}^{(p+1) \times d \times k}$, $\boldsymbol E_{to} \in \mathbb{R}^{(p+1) \times d \times k}$ for all nodes, where each row of $\boldsymbol E_{so}$ represents the source embedding of a node, each row of $\boldsymbol E_{to}$ represents the target embedding of a node. Then, we can infer the causal structure between each pair of nodes by multiplying  $\boldsymbol E_{so}$ and $\boldsymbol E_{to}$:
\begin{equation}
\begin{gathered}
    \boldsymbol{W} = \boldsymbol{E}_{so}(t) \boldsymbol{E}^T_{to}(t)\\
  \boldsymbol{A} = [\boldsymbol{E}_{so}(t-1) \boldsymbol{E}^T_{to}(t-1) | \cdots |\boldsymbol{E}_{so}(t-p) \boldsymbol{E}^T_{to}(t-p) ]
\end{gathered}
\end{equation}
where $\boldsymbol W$ is instantaneous dependency, $\boldsymbol A$ is lagged dependency.  During training, $\boldsymbol{E}_{so}$ and $\boldsymbol{E}_{to}$ will be updated automatically to learn the causal structure among different variables. Compared with the CNN-based work in \citep{Sun23c}, the DGPL module is simpler and the learned embedding matrices have better interpretability (see Section \ref{sec:case_study} for more details). Finally, the DGPL enhanced DBN can be formulated as:

\begin{equation}
\begin{aligned}
    X_t = X_t (\boldsymbol E_{so}(t) \boldsymbol E^T_{to}(t) \circ \boldsymbol{M}_{\tau, \omega}(\boldsymbol{p})_{u,v}) + Y \boldsymbol{{A}}_{so} \odot \boldsymbol{{A}}_{to} + Z
\end{aligned}
\label{eq11}
\end{equation}
where $\boldsymbol{{A_{so}}}=[E_{so}(t-1)| \cdots| E_{so}(t-p)] \in \mathbb {R}^{pd \times k}$ is the lagged source embedding matrix, and $\boldsymbol{{A_{to}}}=[E_{to}(t-1)| \cdots| E_{to}(t-p)] \in \mathbb {R}^{pd \times k}$ is the lagged target embedding matrix, ''$\odot$" means that each entry matrix in the two matrices is multiplied.

\subsection{Nonlinear Form and Model Training}
\subsubsection{Nonlinear Form}
In practice, the interactions among variables can be highly nonlinear, increasing the difficulty in modeling. To alleviate this issue, we adopt \textbf{\textit{1D Convolutional Neural Networks} (CNN)}, a classical nonlinear model for temporal data, to capture nonlinear interactions. We replace the 3D convolutional kernel with two consecutive kernels, which are estimates of $\boldsymbol{E}_{so}$ and $\boldsymbol{E}_{to}$. Formally, a convolutional kernel in a 1D CNN is 3D tensor $\mathcal{W} \in \mathbb{R}^{K \times d \times d}$, where $K= p + 1$ is the number of convolution kernels and $d$ is the input feature dimension. Let $X_{t-p:t} \in \mathbb{R}^{B \times T \times d}$ be the input temporal data, then the output is defined as:  
\begin{equation}
\begin{gathered}
   X_t =  \mathcal{W}^{p+1}  \ast ( \mathcal{W}^{p} \ast \cdots (\mathcal{W}^{1} \ast X_{t-p:t})) = \theta^T X_{t-p:t} \\
   s.t. \text{ } \mathcal{W}^{c} =\boldsymbol{E}^c_s(\boldsymbol{E}^c_t)^T, \forall c \in \{1, \cdots, p+1\}
\end{gathered}
\end{equation}
where $\mathcal{W}^{c}$ represents the $c$-th filter, $\theta^T$ represents the weight obtained by combining multiple layers of $\mathcal{W}^{c}$, $\boldsymbol{E}^c_{so}$ is to learn whether the variable $X_j$ is the root cause of other variables in the causal structure, $\boldsymbol{E}^c_{to}$ is to learn whether the variable $X_j$ is affected by other root causes, while $\boldsymbol{E}_{so}$ and $\boldsymbol{E}_{to}$ are denoted as collections of $\boldsymbol{E}^c_{so}$ and $\boldsymbol{E}^c_{to}$. ''$\ast$" is the convolution operation. 

Similar to \citep{Sun23c}, the first layer of each CNN is a 1D convolution layer with $p+1$ kernels, stride equal to 1, and no padding, where the last kernels $p+1$ represent instantaneous dependency.  Finally, the CNN enhanced \textbf{LOCAL} (\textbf{LOCAL-CNN}) can be formulated as:
\begin{equation}
\begin{gathered}
    \hat{X}_{t} = CNN(X_{t-p:t}; \theta^{(q)})\\
    s.t. \text{ } \mathcal{W}^{p+1(q)} := \mathcal{W}^{p+1(q)} \circ \boldsymbol{M}_{\tau, \omega}(\boldsymbol{p}^{(q)})_{u,v})
\end{gathered}
\end{equation}
\subsubsection{Model Training}
In this part, we combine the ACML and DGPL modules with QMLE, then the overall loss function $\mathcal{S}(\boldsymbol{W}, \boldsymbol{A}; X_t)$ becomes:

\begin{equation}
\begin{aligned}
    \mathcal{S}(\boldsymbol{W}, \boldsymbol{A}; X_t) &= \mathcal{S}(\boldsymbol E_{so}, \boldsymbol E_{to}, \boldsymbol{p}; X_t) \\
    &  =\frac{d}{2} \log ( \|\boldsymbol{D}\boldsymbol{S}^T \{\boldsymbol{S}X_t - Y\boldsymbol{A}\}\|^2) - \log|\text{det}(\boldsymbol{S})|\\
    & + \lambda_2 (\mathcal{R}_{sparse}(\boldsymbol{A}) + \mathcal{R}_{sparse}(\boldsymbol{W}))
\end{aligned}
\end{equation}
where $\boldsymbol{S}:= \boldsymbol{I}-\boldsymbol E_{so}(t) \boldsymbol E^T_{to}(t) \circ \boldsymbol{M}_{\tau, \omega}(\boldsymbol{p})_{u,v}$, $A:=\boldsymbol{{A}}_{so} \odot \boldsymbol{{A}}_{to}$, $W :=\boldsymbol E_{so}(t) \boldsymbol E^T_{to}(t) \circ \boldsymbol{M}_{\tau, \omega}(\boldsymbol{p})_{u,v} $. 

For the nonlinear form, the overall objective becomes:
\begin{equation}
\begin{aligned}
    \mathcal{S}(\boldsymbol{W}, \boldsymbol{A}; X_t) &= \mathcal{S}(\boldsymbol E_{so}, \boldsymbol E_{to}, \boldsymbol{p}; X_t) \\
    &  =\frac{d}{2} \log ( \|\boldsymbol{D}\boldsymbol{S}^T \{X_t - CNN(X_{t-p:t}; \theta^{(q)})\}\|^2) \\
    & - \log|\text{det}(\boldsymbol{S})| + \lambda_2 (\mathcal{R}_{sparse}(\boldsymbol{A}) + \mathcal{R}_{sparse}(\boldsymbol{W}))
\end{aligned}
\end{equation}

\section{Experiments}
\label{sec:exp}



\subsection{Experimental Setup}
This section provides an overview of the datasets, hyper-parameters setting, evaluation metrics, and compared baselines.

\subsubsection{Datasets.} 
The performance of \textbf{LOCAL} was evaluated on three datasets. The first dataset is a synthetic dataset designed to test \textbf{LOCAL}'s ability to recover the underlying causal structures. We generated temporal data through a two-step process: \textbf{1)} Sampling contemporaneous and time-lagged matrices using the \textit{Erd\H{o}s-R\'enyi} scheme. We populated the contemporaneous matrix  $\boldsymbol{W}$ with elements uniformly distributed over the ranges $[-2.0, -0.5] \cup [0.5, 2.0]$. For the time-lagged matrices $A_k$, elements were sampled from $[-1.0\alpha, -0.25\alpha] \cup [0.25\alpha, 2.0\alpha]$, where $\alpha=1/ \eta^k, \eta \leq 1, k=1, \cdots, p$; \textbf{2)} Generating time series data that adhered to the sampled weighted graph according to Eq. (\ref{eq1}). The second dataset, NetSim \citep{smith2011network}, is an fMRI dataset designed to facilitate the study of evolving challenges in brain networks. NetSim comprises 28 datasets, from which we selected 17 with the same sequence length. Each selected dataset consists of 50 independent time series recordings, capturing the activity of 5-15 nodes over 200 time steps. The third dataset, CausalTime \citep{cheng2024causaltime}, is a real-world benchmark dataset containing three types of time series from weather, traffic, and healthcare scenarios. The total length of the dataset is 8760 (Air Quality Index subset), 52116 (Traffic subset), and 40000 (Medical subset) with node counts of 36, 20, and 20, respectively.

\subsubsection{Hyper-parameter Settings.}
In this study, each DGPL module included a source embedding matrix $\boldsymbol{E}_{so} \in \mathbb{R}^{(P+1) \times d \times k} $ and a target embedding matrix $\boldsymbol{E}_{to} \in \mathbb{R}^{(P+1) \times d \times k} $. For the ACML module, we employed a $ \boldsymbol{p} $ vector to represent the priority relationship between nodes and exploit the $\omega$-centered Gumbel sigmoid function to sample the orientation matrix derived. The translation parameter $\omega$ was carefully set to 0.01 to ensure optimal performance. The dimension $k$ of the embedding matrix in DGPL was determined as $\frac{2}{5} \times d$. Under this configuration, the parameter amount of the embedding matrix was marginally smaller than $d \times d$, ensuring efficient representation of the data while mitigating computational complexity. Additionally, the value of the $l_1$ penalty term $\lambda_2$ was set to 0.01 to regulate the sparsity of the learned causal structures. For synthetic datasets, the batch size was set to 16, while for real-world datasets, a batch size of 64 was utilized. Training and inference were performed on an NVIDIA GeForce RTX 3090 machine, with an average training time of 20 minutes.

\subsubsection{Evaluation Metrics} We evaluated the performance of our proposed method for learning causal graphs using three main metrics: 1) True Positive Rate (\textbf{TPR}), which measured the proportion of actual positives that were correctly identified by the model; 2) Structural Hamming Distance (\textbf{SHD}), which count the number of discrepancies (such as reversed, missing, or redundant edges) between two DAGs; 3) \textbf{F1 score}, which represented the harmonic mean of precision and recall, effectively balancing the two in the evaluation process. Given that the number of potential non-causal relationships vastly outnumbered true causal relationships in real datasets, we utilized the Area Under the Precision-Recall Curve (\textbf{AUPRC}) and the Area Under the ROC Curve (\textbf{AUROC}) to evaluate \textbf{LOCAL}’s effectiveness in identifying genuine and significant causal relationships.

\subsubsection{Baselines}
\begin{table}[tb!]
  \centering
  \caption{Comparison of different causal discovery methods. The ``Constraint Free" row indicates whether the method imposes constraints on the directed acyclic graph of instantaneous dependencies. The ``Augmented Lagrangian" row specifies whether the method employs a Lagrangian augmentation approach. The ``Non-linearity" row indicates whether nonlinear dependencies can be captured. The ``Mini-batch" row indicates whether the method supports mini-batch training. The ``Contemporaneous" row and the ``Time-lagged" row refer to the consideration of instantaneous dependencies and time-lagged dependencies.}
  {\small
    \begin{tabular}{c|cccc}
    \toprule
    \toprule
    Method & DYNO & NTS-NO  & TECDI & \textbf{LOCAL} \\
    \midrule
    \midrule
    Optimizer & L-BFGS-B & L-BFGS-B & SGD & Adam \\
    Constraint Free& \ding{55} & \ding{55} & \ding{55} & \ding{51}\\
    Augmented Lagrangian & \ding{51} & \ding{51} & \ding{51} & \ding{55}\\
    Non-linearity & \ding{55} & \ding{51} & \ding{51} & \ding{51}\\
    Mini-batch & \ding{55} & \ding{55} & \ding{51} & \ding{51}\\
    Contemporaneous & \ding{51} & \ding{51} & \ding{51} & \ding{51}\\
    Time-lagged & \ding{51} & \ding{51} & \ding{51} & \ding{51}\\
    \bottomrule
    \bottomrule
    
    \end{tabular}}
  \label{tab:difference}%

\end{table}%

To demonstrate the effectiveness of \textbf{LOCAL} in accurately recovering causal structures from Eq.~(\ref{eq1}) using DBN, we compared it with the following state-of-the-art baselines. Table \ref{tab:difference} summarized the difference between previous methods and \textbf{LOCAL}.

\begin{itemize}
    \item \textbf{DYNO} \citep{Pamfil20a}: DYNO was a scalable method for learning the structure of DBN from time-series data to efficiently estimate both contemporaneous and time-lagged relationships among variables. 
    
    \item \textbf{NTS-NO} \citep{Sun23c}: NTS-NO was an extension of the nonlinear CNN version of DYNO and integrated prior knowledge to improve accuracy. 
    
    \item \textbf{TECDI} \citep{Li2023}: TECDI incorporated intervention data into time series analysis to effectively reveal nonlinear temporal causal relationships, and was the current SOTA model. 
\end{itemize}

To assess whether \textbf{LOCAL} can accurately recover causal structures from complex nonlinear fMRI datasets, we further compared it with several competitive methods.

\begin{itemize}
    \item \textbf{NGC} \citep{Tank2021}:  NGC was a neural Granger causal structure learning method that leveraged an LSTM or MLP to predict the future and conduct causal discovery based on input weights.
    \item \textbf{PCMCI} \citep{NIPS2013_47d1e990}: PCMCI was a causal structure learning algorithm based on non-linear independence tests of time series.
\end{itemize}

To evaluate whether the causal structures identified by \textbf{LOCAL} align with expert knowledge on real-world datasets, we further compared it with recent and representative causal discovery methods, as directly reproduced by Cheng \citep{cheng2024causaltime}.

To verify the effectiveness of the quasi-likelihood objective, DGPL module, and ACML module, we introduced three variants of \textbf{LOCAL} as follows:

\begin{itemize}
    \item \textbf{LOCAL} \textbf{w/o} ACML. It removed the ACML module with a causal constraint.
    \item \textbf{LOCAL} \textbf{w/o} DGPL. It removed the DGPL module with a  $d \times d$ causal matrix.
    \item \textbf{LOCAL} \textbf{w/o} QMLE. It removed the quasi-maximum likelihood estimator (QMLE) with the LSE.
\end{itemize}
 
\subsection{Results on synthetic datasets}
\label{Section:simulation}

\begin{figure}[tb!]
\centering
\includegraphics[width=6.5in]{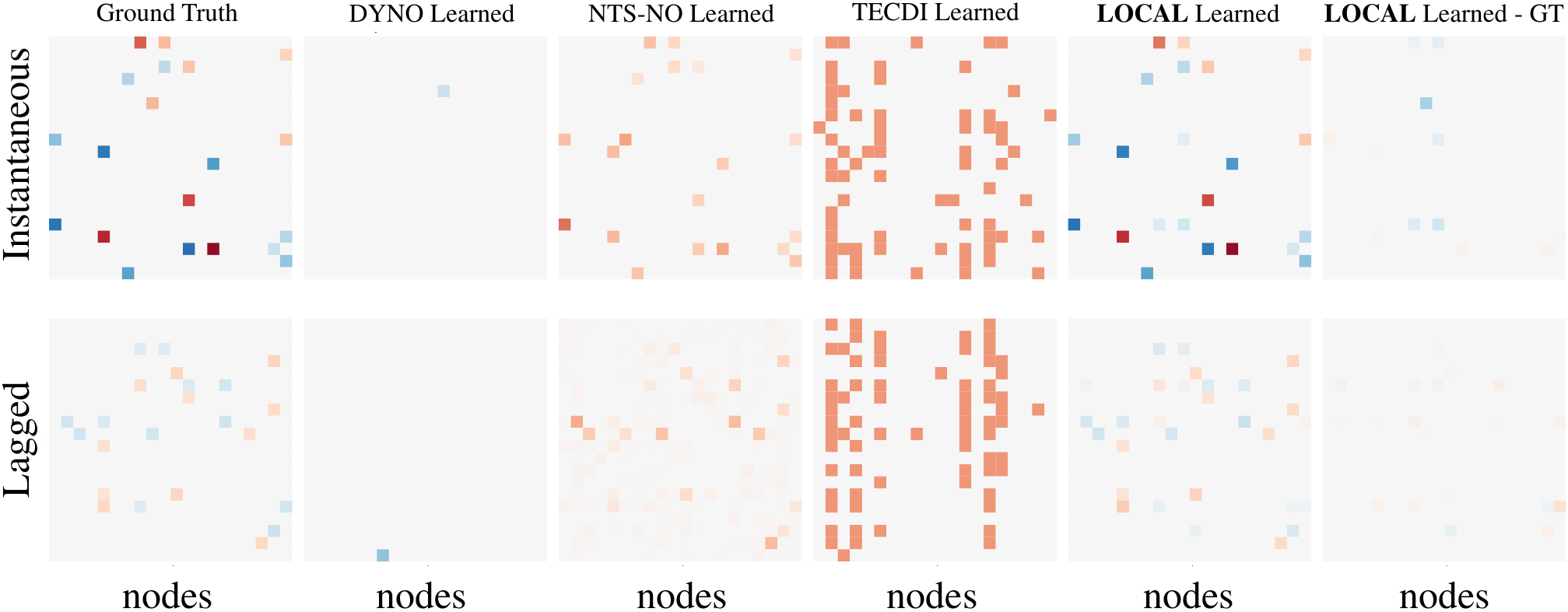}
\DeclareGraphicsExtensions.
\caption{Comparison of dependencies learned by various causal discovery models with the ground truth on a synthetic dataset with $d=20$ nodes, $T=1000$ time steps, and a lag order of $p=1$. The results indicate that the dependencies learned by LOCAL closely match the ground truth for both lagged and instantaneous dependencies.}
\label{Learned}
\end{figure}

We conducted experiments across three synthetic datasets, each respectively containing \{5, 10, 20, 50, 100\} nodes. Each node was accompanied by $p=1$ autoregression terms. Correspondingly, each dataset included \{5, 10, 20, 50, 100\} distinct intervention targets, with each target focused on a different node. Figure \ref{Learned} showed the learned instantaneous matrix $\boldsymbol{W}$ and lagged matrix $\boldsymbol{A}$ on dataset 1 under different algorithms. It could be observed that compared with others, the causal structure learned by \textbf{LOCAL} was almost consistent with the ground truth.

\begin{figure*}[tb!]
\centering
\includegraphics[width=6in]{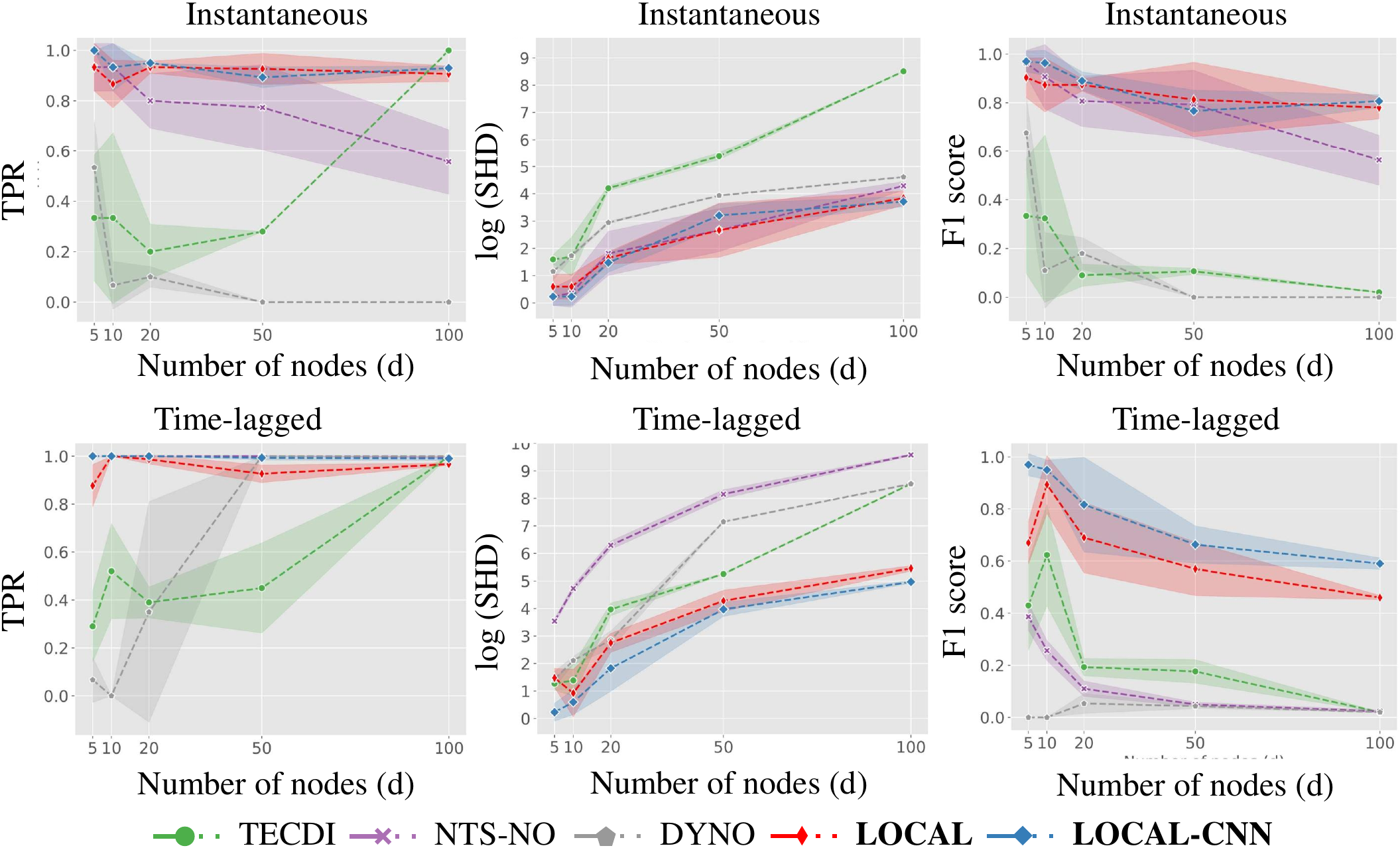}
\caption{Comparison of instantaneous dependencies (upper) and lagged dependencies (lower) across various methods on a synthetic dataset with different numbers of nodes, $d = \{5, 10, 20, 50, 100\}$. The performance of each method is evaluated using three metrics: True Positive Rate (TPR), Structural Hamming Distance (SHD), and F1 score. Each row present the results for these three metrics, assessing both intra-slice and inter-slice graphs. The results indicate that LOCAL consistently performe well across different metrics and node configurations.}
\label{Simulation}
\end{figure*}

\setcounter{table}{1}
\begin{table}[tb!]
  \centering
  \caption{Evaluation results of different causal discovery methods on a synthetic dataset with low-dimensional settings ($d \leq 20$). The performance is assessed using True Positive Rate (TPR), Structural Hamming Distance (SHD), and F1 score. Higher TPR and F1 scores indicate better causal discovery performance, while lower SHD values reflect more accurate graph structures.}
  \resizebox{1.0 \textwidth}{!}{
    \begin{tabular}{cccccccccccc}
    \toprule
    \toprule
    \multirow{2}{*}{Methods} &   \multicolumn{3}{c}{d=5, lag=1}&  & \multicolumn{3}{c}{d=10, lag=1} & &\multicolumn{3}{c}{d=20, lag=1}\\
     \cmidrule{2-4} \cmidrule{6-8} \cmidrule{10-12}
    & TPR $\uparrow$ & SHD $\downarrow$ & F1 score $\uparrow$ & & TPR $\uparrow$ & SHD $\downarrow$ & F1 score $\uparrow$ & &TPR $\uparrow$ & SHD $\downarrow$ & F1 score $\uparrow$ \\
    \midrule
        DYNO \citep{Pamfil20a} & \numstd{0.53}{0.19} & \numstd{2.33}{0.94} & \numstd{0.68}{0.15}  &  & \numstd{0.07}{0.09} & \numstd{4.67}{0.47} & \numstd{0.11}{0.16} &  & \numstd{0.10}{0.04} & \numstd{18.00}{0.82} & \numstd{0.18}{0.07} \\
        NTS-NO \citep{Sun23c} & \numstd{0.93}{0.09} & \numstd{\underline{0.33}}{0.47} & \numstd{\underline{0.96}}{0.05} &  & \numstd{\underline{0.93}}{0.09}& \numstd{\underline{0.67}}{0.94}& \numstd{\underline{0.91}}{0.13}&  & \numstd{0.80}{0.11} & \numstd{17.67}{11.44}& \numstd{0.79}{0.14}\\
        TECDI \citep{Li2023} & \numstd{0.33}{0.25} & \numstd{4.00}{0.82} & \numstd{0.33}{0.24}  &  & \numstd{0.33}{0.34}& \numstd{5.67}{3.40} & \numstd{0.32}{0.34} &  & \numstd{0.20}{0.11} & \numstd{66.00}{5.89} & \numstd{0.09}{0.05} \\
        \rowcolor{mycolor} \textbf{LOCAL}  & \numstd{\underline{0.93}}{0.09} & \numstd{1.00}{0.82}& \numstd{0.90}{0.08}&  & \numstd{0.87}{0.09}& \numstd{1.00}{0.82} & \numstd{0.87}{0.11} &  & \numstd{\underline{0.93}}{0.02}& \numstd{\underline{4.33}}{1.25} & \numstd{\underline{0.87}}{0.69} \\
        \rowcolor{mycolor} \textbf{LOCAL-CNN}  & \numstd{\textbf{1.00}}{0.00} & \numstd{\textbf{0.33}}{0.47} & \numstd{\textbf{0.97}}{0.04} &  & \numstd{\textbf{0.93}}{0.09} & \numstd{\textbf{0.33}}{0.47} & \numstd{\textbf{0.96}}{0.05} & & \numstd{\textbf{0.95}}{0.00} & \numstd{\textbf{3.67}}{1.70} &  \numstd{\textbf{0.89}}{0.04} \\
    \midrule
        DYNO \citep{Pamfil20a} & \numstd{0.07}{0.09} & \numstd{3.33}{1.25} & \numstd{0.00}{0.00} &  & \numstd{0.00}{0.00} & \numstd{7.33}{1.25} & \numstd{0.00}{0.00} &  & \numstd{0.35}{0.46} & \numstd{17.33}{5.19}& \numstd{0.05}{0.04} \\
        NTS-NO \citep{Sun23c} & \numstd{\underline{1.00}}{0.00}& \numstd{33.67}{3.09} & \numstd{0.39}{0.05} &  & \numstd{1.00}{0.00} & \numstd{113.00}{13.36} & \numstd{0.26}{0.04} & & \numstd{\underline{1.00}}{0.00} & \numstd{546.67}{85.06} & \numstd{0.11}{0.03} \\
        TECDI \citep{Li2023} & \numstd{0.29}{0.15} & \numstd{\underline{2.67}}{0.94} & \numstd{0.43}{0.17} &  & \numstd{0.52}{0.20} & \numstd{3.67}{2.49} & \numstd{0.62}{0.19} & & \numstd{0.39}{0.06} & \numstd{53.33}{11.90} & \numstd{0.19}{0.03} \\
         \rowcolor{mycolor} \textbf{LOCAL}  & \numstd{0.88}{0.09} & \numstd{3.67}{1.70}& \numstd{\underline{0.67}}{0.08}&  & \numstd{\underline{1.00}}{0.00}& \numstd{\underline{2.67}}{3.09}& \numstd{\underline{0.89}}{0.11}& & \numstd{0.99}{0.02}& \numstd{\underline{15.67}}{6.02} & \numstd{\underline{0.69}}{0.13}\\
       \rowcolor{mycolor} \textbf{LOCAL-CNN}  & \numstd{\textbf{1.00}}{0.00} & \numstd{\textbf{0.33}}{0.47} & \numstd{\textbf{0.97}}{0.04} &  & \numstd{\textbf{1.00}}{0.00} & \numstd{\textbf{1.00}}{0.82} & \numstd{\textbf{0.95}}{0.04} &  & \numstd{\textbf{1.00}}{0.00} & \numstd{\textbf{8.00}}{7.79} & \numstd{\textbf{0.82}}{0.18}\\
    \bottomrule
    \bottomrule
    \end{tabular}}%
  \label{tab:1}%
\end{table}%

\begin{table}[tb!]
  \centering
  \caption{Evaluation results of different causal discovery methods on a synthetic dataset with high-dimensional settings ($d \geq 50$). The performance is assessed using True Positive Rate (TPR), Structural Hamming Distance (SHD), and F1 score. Higher TPR and F1 scores indicate better causal discovery performance, while lower SHD values reflect more accurate graph structures.}
  \resizebox{1.0 \textwidth}{!}{
    \begin{tabular}{cccccccc}
    \toprule
    \toprule
    \multirow{2}{*}{Methods} &   \multicolumn{3}{c}{d=50, lag=1}&  & \multicolumn{3}{c}{d=100, lag=1} \\
     \cmidrule{2-4} \cmidrule{6-8}
    & TPR $\uparrow$ & SHD $\downarrow$ & F1 score $\uparrow$ & & TPR $\uparrow$ & SHD $\downarrow$ & F1 score $\uparrow$  \\
    \midrule
    \midrule
        DYNO \citep{Pamfil20a} & \numstd{0.00}{0.00} & \numstd{50.00}{0.00} & \numstd{0.00}{0.00} &  & \numstd{0.00}{0.00} & \numstd{100.00}{0.00} & \numstd{0.00}{0.00} \\
        NTS-NO \citep{Sun23c} & \numstd{0.77}{0.17} & \numstd{\textbf{17.67}}{11.44}& \numstd{\underline{0.79}}{0.14}&  & \numstd{0.56}{0.13} & \numstd{72.67}{10.87} & \numstd{0.56}{0.10} \\
        TECDI \citep{Li2023} & \numstd{0.28}{0.00} & \numstd{184.00}{0.00}& \numstd{0.12}{0.00} &  & \numstd{0.22}{0.00} & \numstd{187.00}{0.00} & \numstd{0.12}{0.00} \\
        \rowcolor{mycolor} \textbf{LOCAL}  & \numstd{\textbf{0.93}}{0.06}& \numstd{\underline{22.33}}{21.82} & \numstd{\textbf{0.81}}{0.15} &  & \numstd{\underline{0.91}}{0.03}& \numstd{\underline{47.00}}{11.86}& \numstd{\underline{0.78}}{0.05}\\
        \rowcolor{mycolor} \textbf{LOCAL-CNN}  & \numstd{\underline{0.89}}{0.04}& \numstd{26.00}{11.78}& \numstd{0.77}{0.08}&  & \numstd{\textbf{0.93}}{0.01}& \numstd{\textbf{40.67}}{8.18}& \numstd{\textbf{0.81}}{0.03}\\
    \midrule
        DYNO \citep{Pamfil20a} & \numstd{\textbf{1.00}}{0.00} & \numstd{1272.33}{1.25} & \numstd{0.04}{0.01} &  & \numstd{\textbf{1.00}}{0.00} & \numstd{5048.00}{0.82} & \numstd{0.02}{0.00} \\
        NTS-NO \citep{Sun23c} & \numstd{\underline{1.00}}{0.00} & \numstd{3510.33}{544.48} & \numstd{0.05}{0.01} &  & \numstd{\underline{1.00}}{0.00} & \numstd{14462.00}{425.94} & \numstd{0.02}{0.00} \\
        TECDI \citep{Li2023} & \numstd{0.22}{0.00} & \numstd{187.00}{0.00} & \numstd{0.12}{0.00} &  & \numstd{1.00}{0.00}& \numstd{5049.00}{0.00} & \numstd{0.02}{0.0.00} \\
        \rowcolor{mycolor} \textbf{LOCAL}  & \numstd{0.93}{0.04}& \numstd{\underline{77.00}}{31.63}& \numstd{\underline{0.57}}{0.10} &  & \numstd{0.97}{0.00}& \numstd{\underline{234.33}}{22.51}& \numstd{\underline{0.46}}{0.01}\\
        \rowcolor{mycolor} \textbf{LOCAL-CNN}  & \numstd{0.99}{0.01}& \numstd{\textbf{53.67}}{12.55}& \numstd{\textbf{0.66}}{0.07}&  & \numstd{0.99}{0.01} & \numstd{\textbf{143.00}}{6.38}& \numstd{\textbf{0.59}}{0.02}\\
    \bottomrule
    \bottomrule
    \end{tabular}}%
  \label{tab:2}%
\end{table}%

\begin{figure}[tb!]
\centering
\includegraphics[width=3.2in]{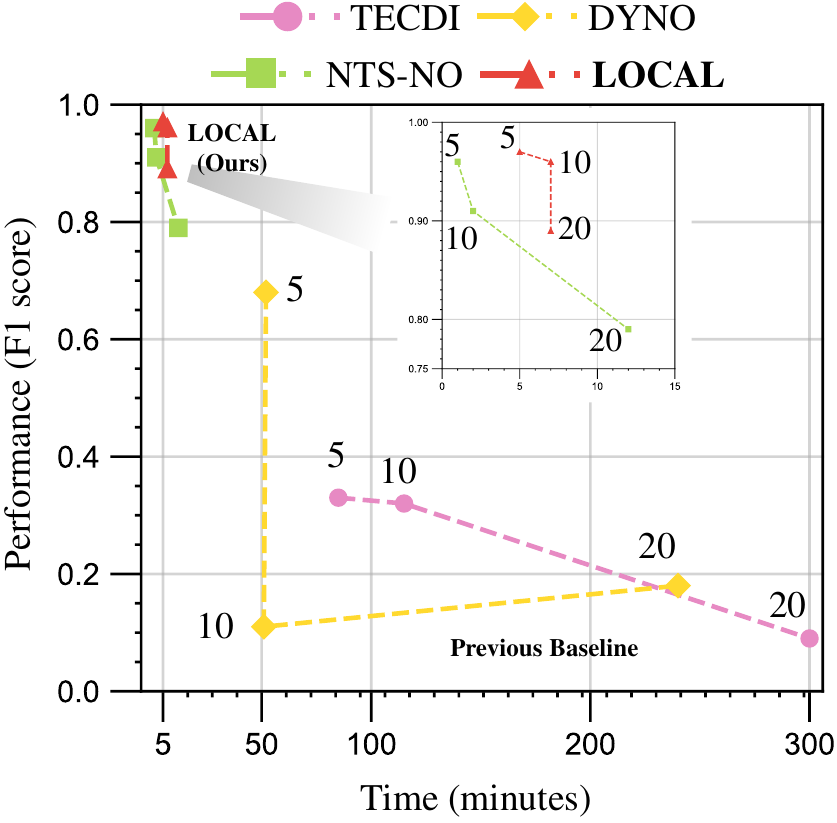}
\DeclareGraphicsExtensions.
\caption{Comparison of total runtime and performance across various causal discovery methods. Each method is evaluated on a synthetic dataset with different numbers of nodes, $d = \{5, 10, 20\}$. The results demonstrate that LOCAL achieve a favorable balance between accuracy and efficiency.}
\label{runningtime}
\end{figure}

Across the synthetic datasets, we conducted three runs of both the baseline methods and \textbf{LOCAL}, averaging the results to obtain final performance indicators. The main results of causal structure learning were shown in Figure \ref{Simulation}. To further exploit the effectiveness of \textbf{LOCAL} under the DBN datasets, we also reported TPR, SHD, and F1 score results in Table \ref{tab:1} and Table \ref{tab:2}.  It was evident that when the variable dimension was high (\emph{e.g.}, $d>50$), \textbf{LOCAL} significantly outperformed the baseline model across all three evaluation metrics (TPR, SHD, and F1 score). Notably, \textbf{LOCAL} exhibited a notable improvement of 226.19\% in TPR. This superior performance could be attributed to the adaptability of the \textbf{LOCAL} decomposition strategy to high-dimensional data, coupled with the inherently sparse and low-rank structure of the simulation data. Regarding computational efficiency, as depicted in Figure \ref{runningtime}, \textbf{LOCAL} demonstrated an average time savings of 97.83\% compared to other score-based algorithms. This efficiency gained stems from \textbf{LOCAL}'s avoidance of matrix exponential operations and its utilization of a first-order optimizer. These computational advantages contributed to the practical scalability of \textbf{LOCAL} in analyzing large-scale datasets.

\subsection{Results on NetSim Datasets}
\setcounter{table}{3}
\begin{table*}[tb!]
  \centering
  \caption{Evaluation results of different causal discovery methods on the NetSim dataset. The performance is assessed using AUROC and AUPRC.}
  \resizebox{1.0 \textwidth}{!}{
    \begin{tabular}{c c c c c c >{\columncolor{mycolor}}c c c c c c c c >{\columncolor{mycolor}}c >{\columncolor{mycolor}}c c}
    \toprule
    \toprule
    \multirow{2}{*}{Dateset} &   \multicolumn{6}{c}{AUPRC}&  & \multicolumn{6}{c}{AUROC}\\
    \cmidrule{2-8} \cmidrule{10-16}
     &  NGC \citep{Tank2021}  & PCMCI \citep{NIPS2013_47d1e990}  & DYNO \citep{Pamfil20a} & NTS-NO \citep{Sun23c} & TECDI \citep{Li2023} & \textbf{LOCAL} & \cellcolor{mycolor}\textbf{LOCAL-{CNN}} & & NGC \citep{Tank2021}  & PCMCI \citep{NIPS2013_47d1e990} & DYNO \citep{Pamfil20a} & NTS-NO \citep{Sun23c} & TECDI \citep{Li2023} & \textbf{LOCAL} & \textbf{LOCAL-{CNN}}\\
    \midrule
    \midrule
    Sim1  & \numstd{0.42}{0.15} & \numstd{0.39}{0.09} &  \numstd{0.41}{0.08}  & \numstd{0.41}{0.10}  & \numstd{0.67}{0.03} &  \numstd{\underline{0.79}}{0.03} & \cellcolor{mycolor}\numstd{\textbf{0.89}}{0.01} & &\numstd{0.65}{0.12} &\numstd{0.64}{0.12} & \numstd{0.73}{0.08} & \numstd{0.64}{0.12} & \numstd{0.67}{0.03} & \numstd{\textbf{0.85}}{0.06} & \numstd{\underline{0.78}}{0.05} \\
    Sim2  & \numstd{0.29}{0.11} & \numstd{0.29}{0.11} &  \numstd{0.33}{0.12} & \numstd{0.24}{0.11} & \numstd{\underline{0.79}}{0.02} & \numstd{0.71}{0.01} & \cellcolor{mycolor}\numstd{\textbf{0.87}}{0.02} & & \numstd{0.68}{0.11} & \numstd{0.69}{0.12} & \numstd{0.81}{0.08} & \numstd{0.50}{0.10} & \numstd{0.79}{0.05} & \numstd{\underline{0.87}}{0.01} & \numstd{\textbf{0.87}}{0.02}\\
    Sim3  & \numstd{0.26}{0.12} & \numstd{0.26}{0.12}  & \numstd{0.32}{0.13} & \numstd{0.16}{0.15} & \numstd{\underline{0.73}}{0.05} & \numstd{0.57}{0.01} & \cellcolor{mycolor}\numstd{\textbf{0.76}}{0.00} & & \numstd{0.72}{0.12} & \numstd{0.73}{0.13} &  \numstd{0.85}{0.07} & \numstd{0.51}{0.08} & \numstd{0.73}{0.06} & \numstd{\underline{0.85}}{0.01} & \numstd{\textbf{0.89}}{0.02}\\
    Sim8  & \numstd{0.40}{0.14} & \numstd{0.36}{0.10}   & \numstd{0.36}{0.08} & \numstd{0.42}{0.08} & \numstd{0.58}{0.04} & \numstd{\underline{0.80}}{0.01} & \cellcolor{mycolor}\numstd{\textbf{0.88}}{0.02} & & \numstd{0.62}{0.12} & \numstd{0.61}{0.11} & \numstd{0.66}{0.10} & \numstd{0.49}{0.10} & \numstd{0.58}{0.03} & \numstd{\underline{0.75}}{0.01} & \numstd{\textbf{0.94}}{0.01}\\
    Sim10 & \numstd{0.42}{0.16} & \numstd{0.40}{0.12} & \numstd{0.38}{0.10} & \numstd{0.46}{0.06} & \numstd{0.71}{0.06} & \numstd{\underline{0.86}}{0.03} & \cellcolor{mycolor}\numstd{\textbf{0.90}}{0.01} & & \numstd{0.65}{0.16} & \numstd{0.66}{0.15} & \numstd{0.69}{0.12} & \numstd{0.58}{0.11} & \numstd{0.71}{0.06} & \numstd{\textbf{0.83}}{0.00} & \numstd{\underline{0.78}}{0.01}\\
    Sim11 & \numstd{0.25}{0.08} & \numstd{0.25}{0.07} & \numstd{0.26}{0.04} & \numstd{0.21}{0.10} & \numstd{\textbf{0.74}}{0.04} & \numstd{0.53}{0.01} & \cellcolor{mycolor}\numstd{\underline{0.68}}{0.02} & & \numstd{0.67}{0.09} & \numstd{0.68}{0.10} & \numstd{0.77}{0.04} & \numstd{0.47}{0.10} & \numstd{0.74}{0.08} & \numstd{\textbf{0.83}}{0.03} & \numstd{\underline{0.77}}{0.02}\\
    Sim12 & \numstd{0.28}{0.11} & \numstd{0.29}{0.11} & \numstd{0.36}{0.08} & \numstd{0.21}{0.05} & \numstd{\underline{0.79}}{0.05} & \numstd{0.70}{0.02} & \cellcolor{mycolor}\numstd{\textbf{0.87}}{0.01} & & \numstd{0.68}{0.12} & \numstd{0.70}{0.13} & \numstd{0.83}{0.05} & \numstd{0.47}{0.11} & \numstd{0.79}{0.04} & \numstd{\underline{0.87}}{0.05} & \numstd{\textbf{0.88}}{0.02}\\
    Sim13 & \numstd{0.47}{0.11} & \numstd{0.47}{0.10} & \numstd{0.47}{0.05} & \numstd{0.53}{0.07} & \numstd{0.68}{0.07} & \numstd{\textbf{0.77}}{0.02} & \cellcolor{mycolor}\numstd{\underline{0.74}}{0.02} & & \numstd{0.59}{0.12} & \numstd{0.59}{0.12} & \numstd{0.66}{0.08} & \numstd{0.43}{0.10} & \numstd{0.68}{0.07} & \numstd{\underline{0.76}}{0.00} & \numstd{\textbf{0.82}}{0.02}\\
    Sim14 & \numstd{0.41}{0.13} & \numstd{0.38}{0.09}  & \numstd{0.41}{0.08} & \numstd{0.42}{0.08} & \numstd{0.67}{0.03} & \numstd{\underline{0.72}}{0.05} & \cellcolor{mycolor}\numstd{\textbf{0.92}}{0.04} & & \numstd{0.65}{0.13} & \numstd{0.64}{0.11} & \numstd{0.74}{0.08} & \numstd{0.39}{0.12} & \numstd{0.67}{0.10} & \numstd{\underline{0.76}}{0.01} & \numstd{\textbf{0.86}}{0.02}\\
    Sim15 & \numstd{0.47}{0.20} & \numstd{0.41}{0.10} & \numstd{0.38}{0.07} & \numstd{0.42}{0.08} & \numstd{0.72}{0.05} & \numstd{\underline{0.80}}{0.03} & \cellcolor{mycolor}\numstd{\textbf{0.86}}{0.02} & & \numstd{0.68}{0.16} & \numstd{0.66}{0.12} & \numstd{0.68}{0.07} & \numstd{0.46}{0.08} & \numstd{0.72}{0.05} & \numstd{\textbf{0.80}}{0.01} & \numstd{\underline{0.79}}{0.02}\\
    Sim16 & \numstd{0.46}{0.10}  & \numstd{0.44}{0.06} & \numstd{0.44}{0.05} & \numstd{0.50}{0.06} & \numstd{0.64}{0.06} & \numstd{\underline{0.73}}{0.00} & \cellcolor{mycolor}\numstd{\textbf{0.91}}{0.02} & & \numstd{0.59}{0.11} & \numstd{0.59}{0.09} & \numstd{0.64}{0.07} & \numstd{0.45}{0.15} & \numstd{0.64}{0.11} & \numstd{\underline{0.73}}{0.02} & \numstd{\textbf{0.93}}{0.03}\\
    Sim17 & \numstd{0.40}{0.19} & \numstd{0.35}{0.10} & \numstd{0.39}{0.09} & \numstd{0.22}{0.12}  & \numstd{\underline{0.86}}{0.05} & \numstd{0.76}{0.03} & \cellcolor{mycolor}\numstd{\textbf{0.90}}{0.01} & & \numstd{0.77}{0.13} & \numstd{0.76}{0.13} & \numstd{\underline{0.87}}{0.05} & \numstd{0.52}{0.12} & \numstd{0.86}{0.05} & \numstd{0.83}{0.01} & \numstd{\textbf{0.87}}{0.02}\\
    Sim18 & \numstd{0.42}{0.16}& \numstd{0.40}{0.11} & \numstd{0.42}{0.07} & \numstd{0.41}{0.06}  & \numstd{0.68}{0.08} & \numstd{\underline{0.75}}{0.01} & \cellcolor{mycolor}\numstd{\textbf{0.89}}{0.02} & & \numstd{0.65}{0.16} & \numstd{0.64}{0.14} & \numstd{0.74}{0.08} & \numstd{0.45}{0.12} & \numstd{0.68}{0.01} & \numstd{\textbf{0.82}}{0.03} & \numstd{\underline{0.80}}{0.00}\\
    Sim21 & \numstd{0.41}{0.14} & \numstd{0.38}{0.09}  & \numstd{0.42}{0.08} & \numstd{0.43}{0.09} & \numstd{0.68}{0.05} & \numstd{\underline{0.79}}{0.01} & \cellcolor{mycolor}\numstd{\textbf{0.88}}{0.02} & & \numstd{0.64}{0.13} & \numstd{0.63}{0.12} & \numstd{0.74}{0.08} & \numstd{0.44}{0.08} & \numstd{0.68}{0.02} & \numstd{\underline{0.80}}{0.04} & \numstd{\textbf{0.83}}{0.03} \\
    Sim22 & \numstd{0.35}{0.09} & \numstd{0.37}{0.08} & \numstd{0.38}{0.06} & \numstd{0.45}{0.06} & \numstd{\textbf{0.83}}{0.02} & \numstd{0.69}{0.03} & \cellcolor{mycolor}\numstd{\underline{0.80}}{0.01} & & \numstd{0.58}{0.13} & \numstd{0.61}{0.12} & \numstd{0.66}{0.07} & \numstd{0.48}{0.09} & \numstd{\textbf{0.83}}{0.10} & \numstd{\underline{0.76}}{0.01} & \numstd{0.72}{0.02} \\
    Sim23 & \numstd{0.45}{0.20} & \numstd{0.41}{0.21} & \numstd{0.35}{0.06} & \numstd{0.41}{0.05} & \numstd{\textbf{0.78}}{0.05} & \numstd{0.64}{0.01} & \cellcolor{mycolor}\numstd{\underline{0.64}}{0.00} & & \numstd{0.67}{0.15} & \numstd{0.65}{0.11} & \numstd{0.64}{0.06} & \numstd{0.43}{0.07} & \numstd{\textbf{0.78}}{0.01} & \numstd{\underline{0.75}}{0.02} & \numstd{0.67}{0.03} \\
    Sim24 & \numstd{0.34}{0.11} & \numstd{0.35}{0.11} & \numstd{0.31}{0.07} & \numstd{0.47}{0.08} & \numstd{\textbf{0.79}}{0.05} & \numstd{\underline{0.65}}{0.02} & \cellcolor{mycolor}\numstd{0.57}{0.02} &  & \numstd{0.55}{0.13} & \numstd{0.57}{0.12} & \numstd{0.53}{0.01} & \numstd{0.47}{0.09} & \numstd{\textbf{0.79}}{0.06} & \numstd{0.69}{0.04} & \numstd{\underline{0.71}}{0.01} \\
    Average & \numstd{0.38}{0.12}  & \numstd{0.36}{0.10} & \numstd{0.38}{0.08} & \numstd{0.37}{0.10} & \numstd{\underline{0.73}}{0.04} & \numstd{0.72}{0.02} & \cellcolor{mycolor}\numstd{\textbf{0.82}}{0.02}& & \numstd{0.65}{0.12} & \numstd{0.65}{0.12} & \numstd{0.72}{0.08}  & \numstd{0.46}{0.07} & \numstd{0.73}{0.04} & \numstd{\underline{0.80}}{0.02} & \numstd{\textbf{0.82}}{0.02}\\
    \bottomrule
    \bottomrule
    \end{tabular}}%
  \label{tab:3}%
\end{table*}%

\setcounter{table}{4}
\begin{table}[tb!]
    \centering
    \caption{F test results comparing the previous baseline methods (\emph{e.g.}, TECDI) with our proposed methods (LOCAL and LOCAL-CNN). ''$F$'' represents the test statistic, while ''$p$" denotes the probability of obtaining the observed results under the null hypothesis. A $p$-value below the significance level (\emph{e.g.}, 0.05) indicates significant differences between the compared algorithms.}
    \begin{tabular}{ccc}
        \hline
        Method comparison & $F$ & $p$ \\
        \hline
        TECDI vs. \textbf{LOCAL} & 11.87 & 0.0015 \\
        TECDI vs. \textbf{LOCAL-CNN} & 14.53 & 0.0006 \\
        \hline
    \end{tabular}%
  \label{tab:F}%
\end{table}%

\setcounter{table}{5}
\begin{table}[tb!]
  \centering
  \caption{Evaluation results of different causal discovery methods on the CausalTime dataset. The performance is assessed using AUROC and AUPRC.}
  \resizebox{1.0 \textwidth}{!}{
    \begin{tabular}{cccccccc}
    \toprule
    \toprule
    \multirow{2}{*}{Methods} &   \multicolumn{3}{c}{AUROC}&  & \multicolumn{3}{c}{AUPRC}\\
     \cmidrule{2-4} \cmidrule{6-8}
    & AQI & Traffic & Medical & & AQI & Traffic & Medical \\
    \midrule
    \midrule
        GC \citep{granger1969investigating} & \numstd{0.45}{0.04} & \numstd{0.42}{0.03} & \numstd{0.57}{0.03} &  & \numstd{0.63}{0.02} & \numstd{0.28}{0.00} & \numstd{0.42}{0.03} \\
        SVAR \citep{Sun23c} & \numstd{0.62}{0.04} & \numstd{0.63}{0.00} & \numstd{0.71}{0.02} &  & \numstd{0.79}{0.02} & \numstd{0.58}{0.00} & \numstd{0.68}{0.04} \\
        NTS-NO \citep{Sun23c}  & \numstd{0.57}{0.02} & \numstd{0.63}{0.03} & \numstd{0.71}{0.02}  &  & \numstd{0.71}{0.02} & \numstd{0.58}{0.05} & \numstd{0.46}{0.02} \\
        PCMCI \citep{NIPS2013_47d1e990}  & \numstd{0.53}{0.07} & \numstd{0.54}{0.07} & \numstd{0.70}{0.01} &  & \numstd{0.67}{0.04} & \numstd{0.35}{0.06} & \numstd{0.51}{0.02} \\
        Rhino \citep{gong2023rhino} & \numstd{0.67}{0.10} & \numstd{{0.63}}{0.02} & \numstd{0.65}{0.02} &  & \numstd{0.76}{0.08} & \numstd{0.38}{0.01} & \numstd{0.49}{0.03} \\
        CUTS \citep{yuxiao2023cuts} & \numstd{0.60}{0.00} & \numstd{0.62}{0.02} & \numstd{0.37}{0.03} &  & \numstd{0.51}{0.04} & \numstd{0.15}{0.02} & \numstd{0.15}{0.00} \\
        CUTS+ \citep{Cheng_2024}  & \numstd{\textbf{0.89}}{0.02} & \numstd{0.62}{0.07} & \numstd{\textbf{0.82}}{0.02} &  & \numstd{\underline{0.80}}{0.08} & \numstd{\underline{0.64}}{0.12} & \numstd{0.55}{0.13} \\
        NGC \citep{Tank2021}  & \numstd{0.72}{0.01} & \numstd{0.60}{0.01} & \numstd{0.57}{0.01} &  & \numstd{0.72}{0.01} & \numstd{0.36}{0.05} & \numstd{0.46}{0.01} \\
        NGM \citep{bellot2022neural} & \numstd{0.67}{0.02} & \numstd{0.47}{0.01} & \numstd{0.56}{0.02} &  & \numstd{0.48}{0.02} & \numstd{0.28}{0.01} & \numstd{0.47}{0.02} \\
        LCCM \citep{brouwer2021latent} & \numstd{\underline{0.86}}{0.07} & \numstd{0.55}{0.03} & \numstd{0.80}{0.02} &  & \numstd{\textbf{0.93}}{0.02} & \numstd{0.59}{0.05} & \numstd{\textbf{0.76}}{0.02} \\
        eSRU \citep{Khanna2020Economy}  & \numstd{0.83}{0.03} & \numstd{0.60}{0.02} & \numstd{0.76}{0.04} &  & \numstd{0.72}{0.03} & \numstd{0.49}{0.03} & \numstd{\underline{0.74}}{0.06} \\
        SCGL \citep{10.1145/3357384.3357864}  & \numstd{0.49}{0.05} & \numstd{0.59}{0.06} & \numstd{0.50}{0.02} &  & \numstd{0.36}{0.03} & \numstd{0.45}{0.03} & \numstd{0.48}{0.02} \\
        TCDF \citep{make1010019} & \numstd{0.41}{0.02} & \numstd{0.50}{0.00} & \numstd{0.63}{0.04} &  & \numstd{0.65}{0.01} & \numstd{0.36}{0.00} & \numstd{0.55}{0.03} \\
        \rowcolor{mycolor} \textbf{LOCAL}  & \numstd{0.85}{0.02} & \numstd{\textbf{0.89}}{0.01} & \numstd{\underline{0.81}}{0.03} &   & \numstd{0.67}{0.00}  & \numstd{\textbf{0.68}}{0.01}  &  \numstd{0.73}{0.03} \\
        \rowcolor{mycolor} \textbf{LOCAL-CNN}  & \numstd{0.85}{0.01} & \numstd{\underline{0.83}}{0.00} & \numstd{0.74}{0.02} &   & \numstd{0.70}{0.01}  & \numstd{0.62}{0.01}  &  \numstd{0.67}{0.01} \\
    \bottomrule
    \bottomrule
    \end{tabular}}%
  \label{tab:4}%
\end{table}%

\setcounter{table}{6}
\begin{table*}[!htbp]
  \caption{Ablation study on the components of LOCAL using a synthetic dataset with $d = \{20, 50, 100\}$ nodes, $T = 1000$ timesteps, and $p$-order = 1. The results demonstrate that each component contributes to improved performance of LOCAL.}
  \label{tab:freq}
  \resizebox{1.0 \textwidth}{!}{
  \begin{tabular}{c|ccc|ccc|ccc}
  \toprule
  \toprule
    Dataset      & \multicolumn{3}{c|}{d=20}     & \multicolumn{3}{c|}{d=50}     & \multicolumn{3}{c}{d=100} \\
    \midrule
    \midrule
    Metric & tpr $\uparrow$  & shd $\downarrow$  & f1$\uparrow$     & tpr $\uparrow$  & shd $\downarrow$  & f1$\uparrow$     & tpr $\uparrow$  & shd $\downarrow$ & f1$\uparrow$  \\
    \midrule
    \midrule
    w/o DGPL & \textbf{0.95}/\textbf{1.00} & 6.33/\underline{12.33} & \underline{0.84}/\underline{0.77} & 0.86/\textbf{1.00} & \underline{42.00}/151.33 & \underline{0.66}/0.43 & \underline{0.91}/\textbf{1.00} & \underline{58.67}/565.33 & \underline{0.75}/0.26 \\
    w/o ACML & 0.75/0.90 & 18.67/21.00 & 0.58/0.61 & \underline{0.92}/0.88 & 52.00/\textbf{74.67} & 0.63/\textbf{0.58} & 0.87/0.89 & 66.00/\textbf{150.00} & 0.70/\textbf{0.54}\\
    w/o QMLE & \underline{0.95}/\underline{1.00} & \underline{6.00}/\textbf{9.67} & 0.84/\textbf{0.82} & 0.84/0.91 & 53.00/116.67 & 0.60/0.47 & 0.89/0.96 & 62.33/237.33 & 0.73/0.44\\
    \rowcolor{mycolor} {\bfseries LOCAL} & 0.93/0.99 & \textbf{4.33}/15.67 & \textbf{0.87}/0.69 & \textbf{0.93}/\underline{0.93} & \textbf{22.33}/\underline{77.00} & \textbf{0.81}/\underline{0.57} & \textbf{0.91}/\underline{0.97} & \textbf{47.00}/\underline{234.33} & \textbf{0.78}/\underline{0.46}\\
    \bottomrule
    \bottomrule
  \end{tabular}}%
  \label{tab:5}%
\end{table*}

We conducted the NetSim experiment, which depicted a dataset characterized by high complexity and nonlinear attributes. We showed AUROC and AUPRC results in Table \ref{tab:3}.  Notably, \textbf{LOCAL} demonstrated significant superiority over the baseline method in both linear and nonlinear scenarios, showcasing an average increase of 4.11\% in AUPRC (12.33\% for \textbf{LOCAL-CNN}) and an average increase of 9.59\% in AUROC (12.33\% for \textbf{LOCAL-CNN}) while saving 66.63\% of time. It was worth mentioning that, due to the ground truth of NetSim not adhering to the low-rank assumption, we slightly increased the embedding dimension to $k=20$ for \textbf{LOCAL-CNN}. This adjustment enabled \textbf{LOCAL-CNN} to effectively captured the intricate relationships within the dataset, contributing to its enhanced performance compared to the baseline method.

We presented the F test results in Table \ref{tab:F}. The findings indicated that both \textbf{LOCAL} and \textbf{LOCAL-CNN} significantly outperformed TECDI. Specifically, the test statistic ($F$) values of 11.87 and 14.53, with corresponding $p$-values of 0.0015 and 0.0006, provided strong evidence against the null hypothesis of no difference between the methods. As both $p$-values are well below the standard significance threshold (0.05), we concluded that the proposed methods demonstrated superior performance compared to TECDI.
\subsection{Results on CausalTime Datasets}
We employed a real-world dataset CausalTime to evaluate the performance of \textbf{LOCAL}. A detailed comparison was presented in Table \ref{tab:4}. We observed that: (1) Among all the compared methods, \textbf{LOCAL} performed best in the Traffic subset and second-best in Medical, which demonstrated the effectiveness of the proposed method. (2) \textbf{LOCAL} achieved the average best performance, which was attributed to the helpful role of explicitly instantaneous and lagged causality information. (3) The linear implementation of \textbf{LOCAL} was more suitable for the CausalTime dataset than the nonlinear version, one possible reason was that the causal relationship between the data appeared linear.
\subsection{Ablation Study}
\begin{figure}[tb!]
\centering
\includegraphics[width=5in]{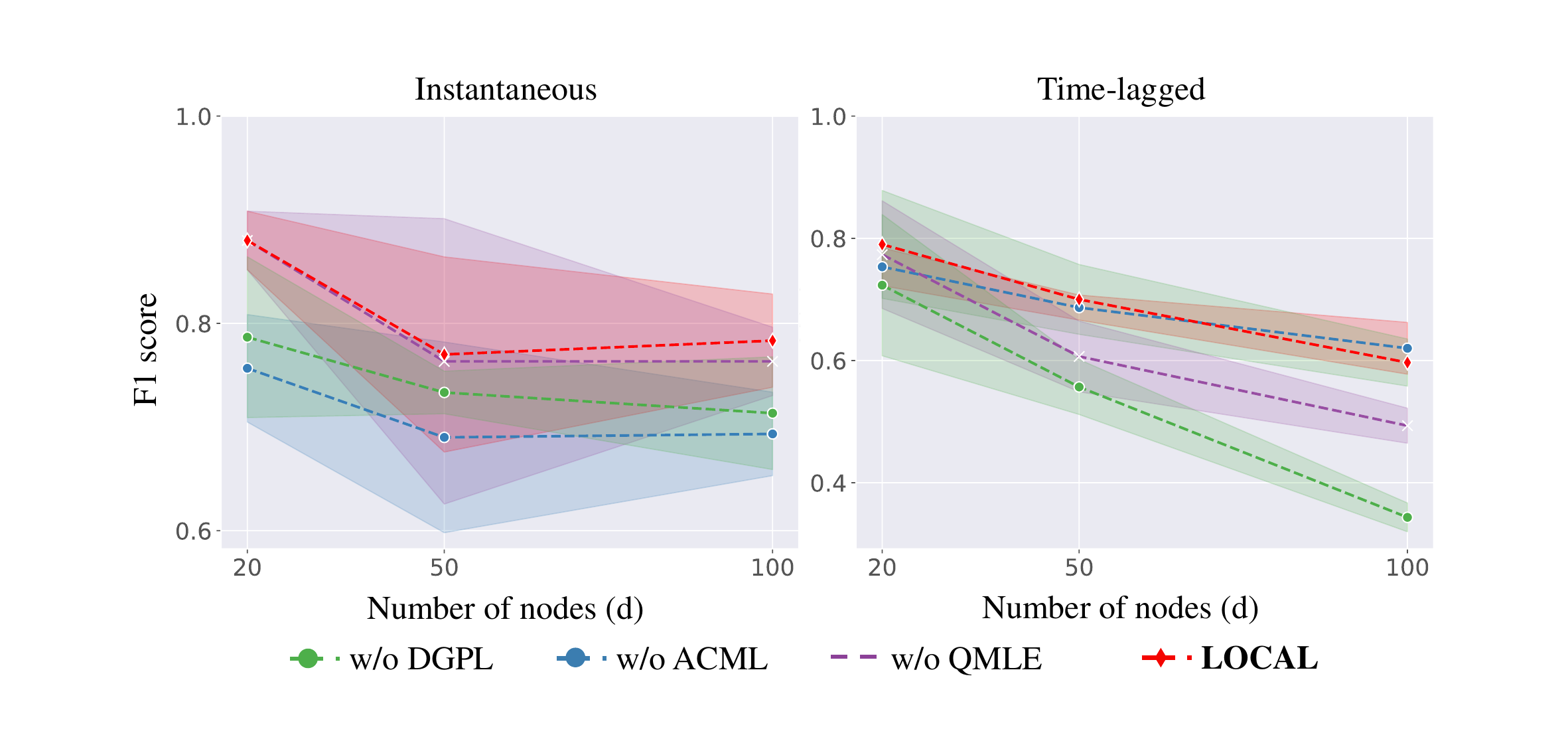}
\DeclareGraphicsExtensions.
\caption{Ablation study of different components in LOCAL. Each component contribute to the learning of instantaneous dependencies (left) and lagged dependencies (right) on a synthetic dataset with $d = \{20, 50, 100\}$ nodes, $T = 1000$ time steps, and a lag order of $p = 1$. The results highlight the impact of removing individual components (DGPL, ACML, and QMLE) on model performance.
}
\label{Ablation}
\end{figure}

We compared \textbf{LOCAL} with three ablation methods. Initially, as presented in Table \ref{tab:freq}, causal relationships were extracted solely from a $d \times d$ matrix. Subsequently, we enhanced the basic low-rank structure by incorporating interventional data and an Asymptotic Causal Mask, leading to substantial performance improvements. Finally, the optimal performance was achieved when all components were integrated into the \textbf{LOCAL} model. This progressive integration of components was illustrated in Figure \ref{Ablation}, highlighting the incremental enhancements in performance as each component was added to the model.

\begin{figure}[!htbp]
\centering
\includegraphics[width=3.2in]{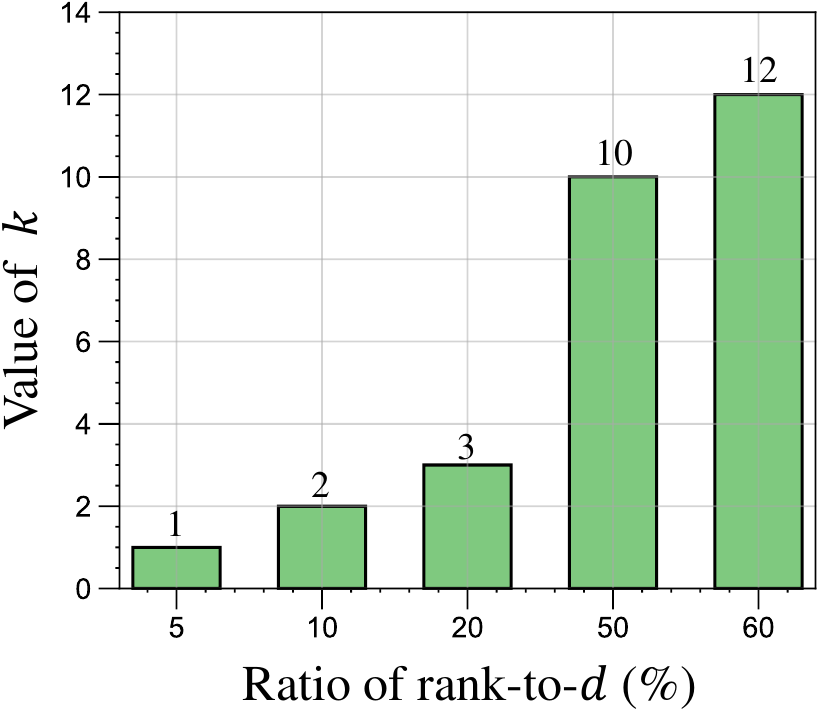}
\DeclareGraphicsExtensions.
\caption{Minimum number of latent dimensions $k$ required for LOCAL to achieve 90\% accuracy at different rank-to-$d$ ratios. Evaluated on a synthetic dataset with $d = \{20, 50, 100\}$ nodes, $T = 1000$ time steps, and a lag order of $p = 1$, the results show that as the rank-to-$d$ ratio increases (5\%, 10\%, 20\%, 50\%, 60\%), the required latent dimension $k$ also increase.}
\label{minrank}
\end{figure}

Figure \ref{minrank} illustrated that to achieve at least 90\% accuracy (using TPR as the evaluation metric), the minimum dimension $k$ increased as the rank-to-$d$ ratio rised, and was approximately proportional to the true rank of the graph structure. This trend might be linked to the inherent data structure: lower matrix ranks often corresponded to graphs with significant centralization. As $k$ increased, graph complexity also rised, necessitating a higher dimension to capture the data's complexity and diversity effectively. This finding aided in selecting an optimal $k$ to maintain accuracy. When the rank was unknown, choosing $k$ to be slightly less than 0.5 times $d$ could balance performance and computational efficiency.

\subsection{Discussion: What do source embedding and target embedding learn respectively?}
\label{sec:case_study}
\begin{figure}[tb!]
\centering
\includegraphics[width=5.0in]{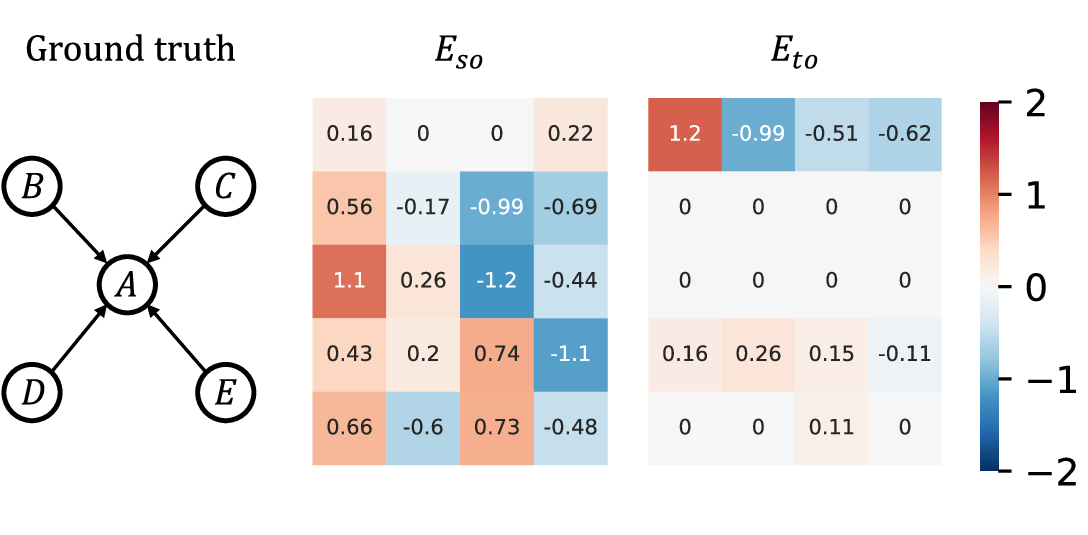}
\DeclareGraphicsExtensions.
\caption{Illustration of the node source embedding matrix ($E_{so}$) and target embedding matrix ($E_{to}$) in a dynamic Bayesian network (DBN) with $d = 5$ nodes. The leftmost graph represent the ground truth structure, while the matrices visualize the learned embeddings.}
\label{LowRank}
\end{figure}

This section visualized the learned source embedding and target embedding in a synthetic dataset with $d=5$. We illustrated the ground truth instantaneous causal graph, source embedding $\boldsymbol{E}_{so}(t)$, and target embedding $\boldsymbol{E}_{to}(t)$ in Figure \ref{LowRank}. As we could see, the $d \times d$-dimensional causal matrix $\boldsymbol{W}$ was decomposed into $\boldsymbol{E}_{so}$ and $\boldsymbol{E}_{to}$, where each row of the source embedding $\boldsymbol{E}_{so}$ corresponded to the variable embedding result in $\boldsymbol{W}$ as parents. Each row of target embedding $\boldsymbol{E}_{to}$ corresponded to the variable embedding result of children. This was exactly why we named it source and target embedding.

\section{Conclusion}
\label{sec:conclusion}
In this work, we introduce an easy-to-implement variant to the dynamic causal discovery family, LOCAL, particularly tailored for capturing nonlinear, high-dimension dynamic systems rapidly. Specifically, it estimates causal structure and connection strength through a quasi-maximum likelihood-based score function, Asymptotic Causal Mask Learning (ACML) and Dynamic Graph Parameter Learning (DGPL) modules respectively. We evaluate our approach on both synthetic and public benchmark datasets, LOCAL has presented advantages over previous state-of-the-art approaches. In summary, LOCAL offers an efficient and flexible solution for dynamic causal discovery in time series, demonstrating strong potential for enhancing both efficiency and performance.

\section{Limitation}
\label{sec:limitation}
While LOCAL demonstrates strong performance across various tasks and datasets, there remain opportunities for further exploration. The DGPL module leverages low-rank matrix factorization to reduce the number of parameters. However, if the true causal structure exhibits a high rank (\emph{e.g.}, densely connected networks), this decomposition may introduce biases in estimating causal strength. Additionally, while our evaluation provides a solid foundation, further research is needed to explore the performance of LOCAL on datasets with a larger number of nodes and longer lagged dependencies. In future work, deep time series modeling, adaptive rank optimization, and non-stationary analysis can be integrated to better capture the causal dynamics of complex real-world systems.

\section*{Acknowledgements}
This work was supported by the National Key Research and Development Program of China under Grant (No.2024YFB3312200), and by the Beijing Natural Science Foundation under Grant (No.L231005).

\section*{Ethical Statement}
1. Articles do not rely on clinical trials.

2. The authors declare that they have no conflict of interest.

3. All submitted manuscripts containing research which does not involves human participants and/or animal experimentation.

\bibliographystyle{cas-model2-names}
\bibliography{sample-base}

\end{document}